\documentclass{article}
\usepackage{spconf,amsmath,graphicx,hyperref}
\usepackage{multirow, booktabs}
\usepackage{amsfonts}
\usepackage{subcaption}

\newcommand{\ie}{\textit{i.e.}~}
\newcommand{\eg}{\textit{e.g.},~}
\newcommand{\etc}{\textit{etc}}

\title{TENSLORA: TENSOR ALTERNATIVES FOR LOW-RANK ADAPTATION}
\twoauthors
 {\parbox{6cm}{\centering Axel Marmoret, Reda Bensaid\\ \textit{Jonathan Lys, Vincent Gripon}}
 }{IMT Atlantique, Lab-STICC, \\UMR CNRS 6285, F-29238 Brest, France}
 {\parbox{6cm}{\centering Fran\c{c}ois Leduc-Primeau}}{Polytechnique Montréal, Canada}

\begin{document}
%
\maketitle
\begin{abstract}

Low-Rank Adaptation (LoRA) is widely used to efficiently adapt Transformers by adding trainable low-rank matrices to attention projections. While effective, these matrices are considered independent for each attention projection (Query, Key, and Value) and each layer. Recent extensions have considered joint, tensor-based adaptations, but only in limited forms and without a systematic framework. We introduce \textit{TensLoRA}, a unified framework that aggregates LoRA updates into higher-order tensors and models a broad family of tensor-based low-rank adaptations. Our formulation generalizes existing tensor-based methods and enables mode-specific compression rates, allowing parameter budgets to be tailored according to the modality and task. Experiments on vision and language benchmarks reveal that the tensor construction directly impacts performance, sometimes better than standard LoRA under similar parameter counts.

\vspace{-3pt}
\end{abstract}
\begin{keywords}
Parameter-Efficient Fine-Tuning, Tensor Factorization, Transformers
\vspace{-3pt}
\end{keywords}
\section{Introduction}
\label{sec:intro}

The growing adoption of foundation models has advanced performance in vision, language, and beyond. These models are typically pre-trained on large datasets and then adapted to downstream tasks. Full fine-tuning, which updates all parameters, is increasingly impractical due to the scale of modern models. Parameter-efficient fine-tuning (PEFT) methods address this by adapting models with only a small number of trainable parameters, while keeping the original model, called ``backbone,'' frozen. Among them, Low-Rank Adaptation (LoRA)~\cite{hu2022lora} has proven particularly effective by injecting low-rank updates into selected weight matrices, most often the attention projections in Transformers~\cite{vaswani2017attention}. LoRA offers low training cost (few parameters), performance on par with or better than full fine-tuning, and no inference overhead since updates can be merged into the backbone. It is considered one of the most efficient PEFT techniques today~\cite{xu2023parameter, bensaid2025a}. 

Despite its strengths, LoRA applies updates independently across attention projections (Query, Key, and Value) and layers, potentially disregarding correlations along these dimensions and creating undesired redundancy; furthermore, the number of parameters scales linearly with the number of adapted components.
Several tensor-based extensions, such as FacT~\cite{jie2023fact}, LoTR~\cite{bershatsky2024lotr}, LoRTA~\cite{hounie2024lorta}, and CaRA~\cite{veeramachenenicanonical}, attempt to reduce redundancy by aggregating LoRA parameters into higher-order tensors. However, these approaches explore only partially ways to extend LoRA to tensors and provide little guidance on which choices are the most effective.

In this work, we introduce \textit{TensLoRA}, a unified framework that systematically covers tensor-based low-rank adaptations by aggregating attention layers into tensors. TensLoRA captures and extends FacT, LoTR, LoRTA, and CaRA, albeit with differences in scope and implementation, particularly with respect to the types of layers considered. Beyond unification, TensLoRA introduces mode-specific compression rates, allowing parameter budgets to be strategically allocated to exploit non-uniform redundancy across different model dimensions. Our experiments show that performance depends on the tensor construction, with some variants surpassing standard LoRA at comparable parameter counts, though none consistently outperforms LoRA at high compression rates. Code to reproduce experiments can be found on GitHub~\footnote{\url{https://github.com/ax-le/TensLoRA/tree/ICASSP26}}.

The article is structured as follows: Section~\ref{sec:lora_tensor_formalism} introduces LoRA and its tensor alternatives, Section~\ref{sec:tensor_formalism} introduces tensor factorization, and Section~\ref{sec:expes} presents the experimental results.


\section{LoRA as Tensors}
\label{sec:lora_tensor_formalism}

\begin{figure*}[ht]
    \centering
    \begin{subfigure}[t]{0.2\textwidth}
        \centering
        \includegraphics[width=0.48\linewidth]{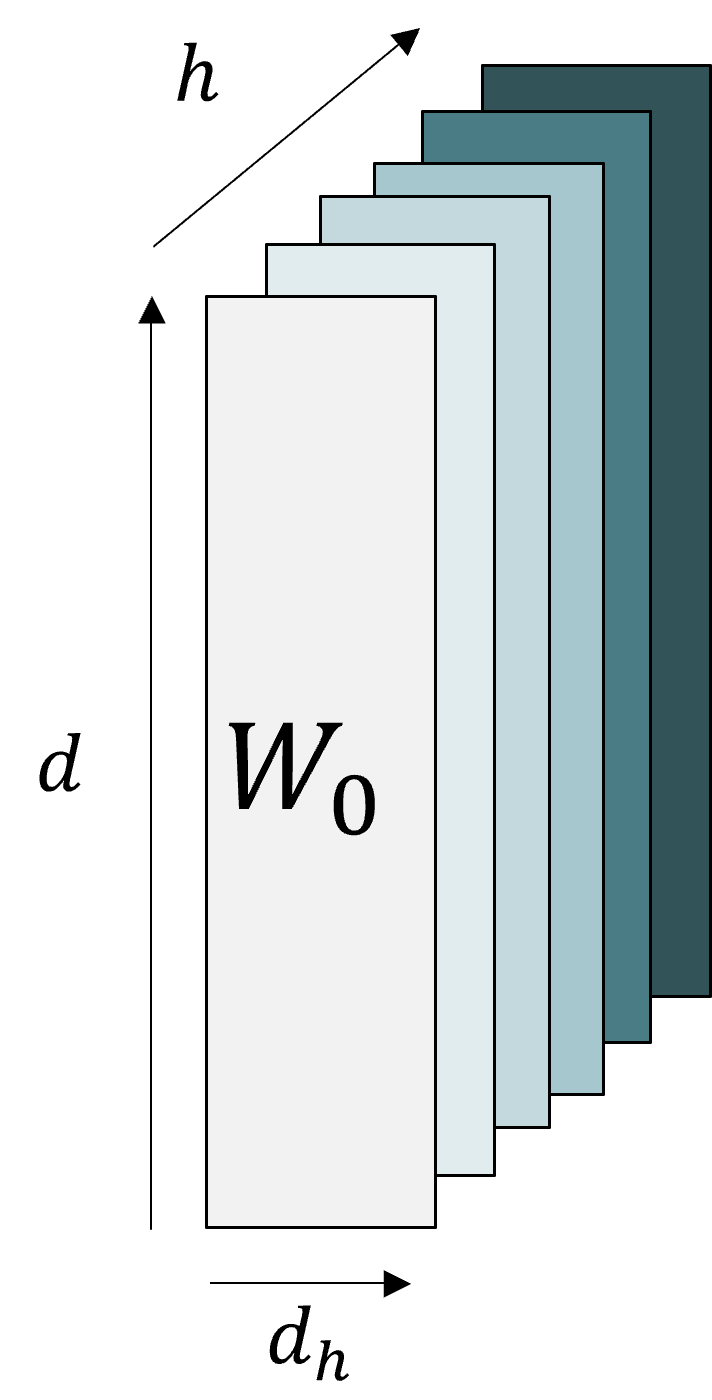}
        \caption{\textbf{Att} tensor}
        \label{fig:att}
    \end{subfigure}
    \begin{subfigure}[t]{0.2\textwidth}
        \centering
        \includegraphics[width=0.8\linewidth]{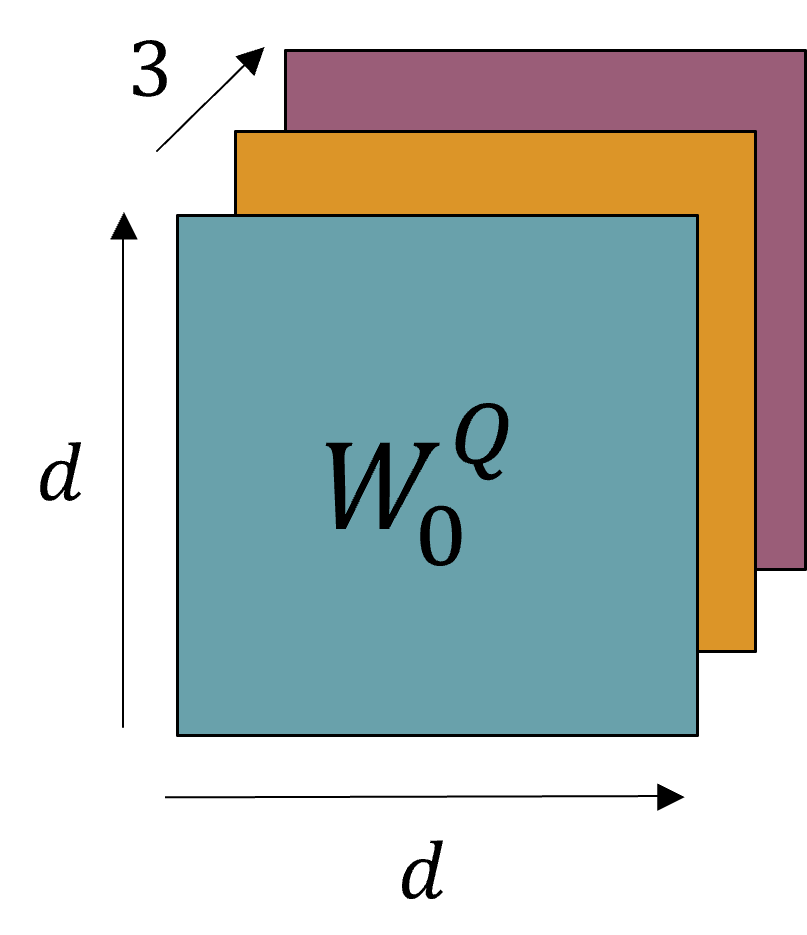}
        \caption{\textbf{QKV} tensor}
            \label{fig:qkv}
    \end{subfigure}
    \begin{subfigure}[t]{0.2\textwidth}
        \centering
        \includegraphics[width=0.9\linewidth]{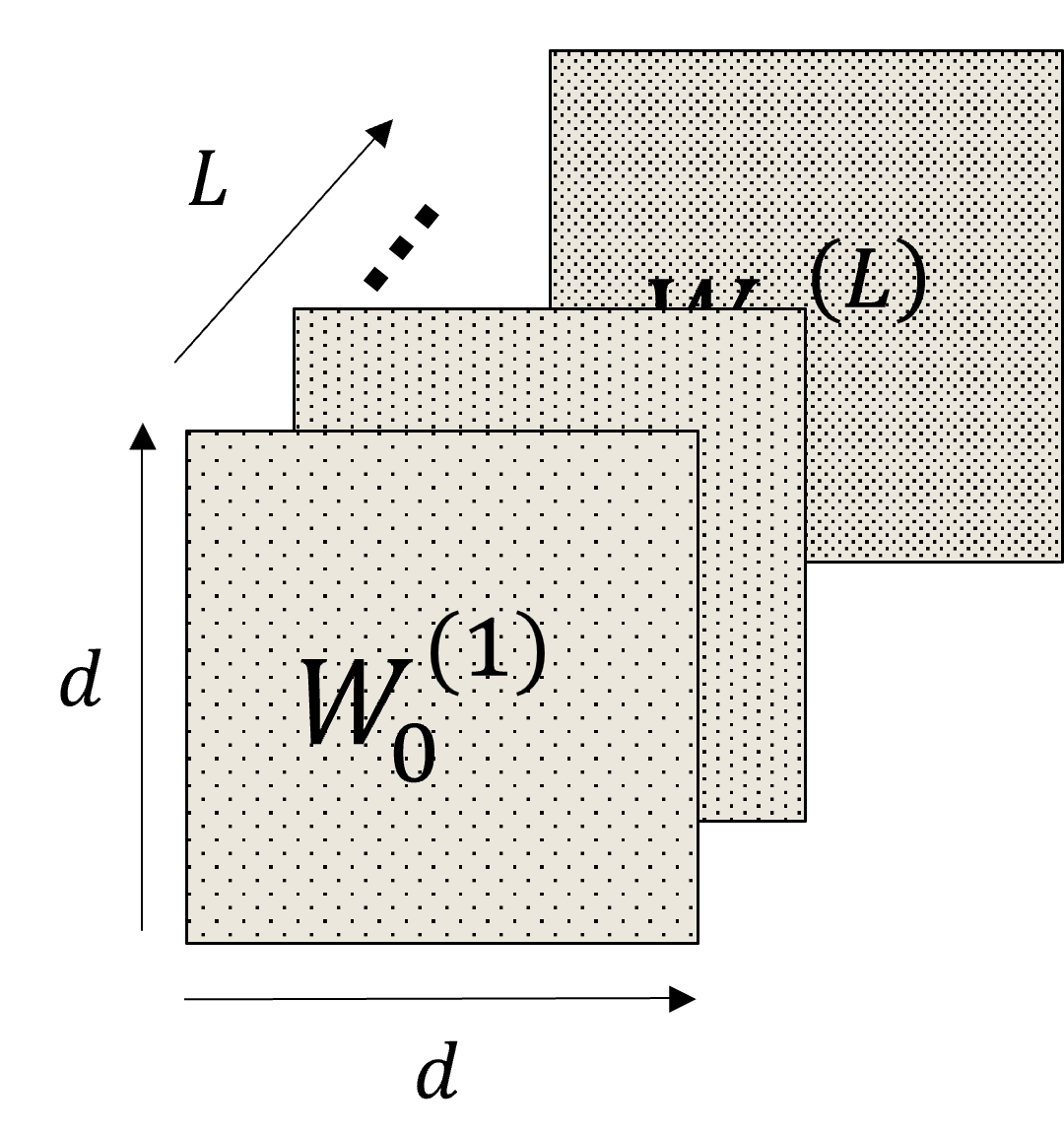}
        \caption{\textbf{Depth} tensor}
        \label{fig:depth}
    \end{subfigure}
    \begin{subfigure}[t]{0.3\textwidth}
        \centering
        \includegraphics[width=\linewidth]{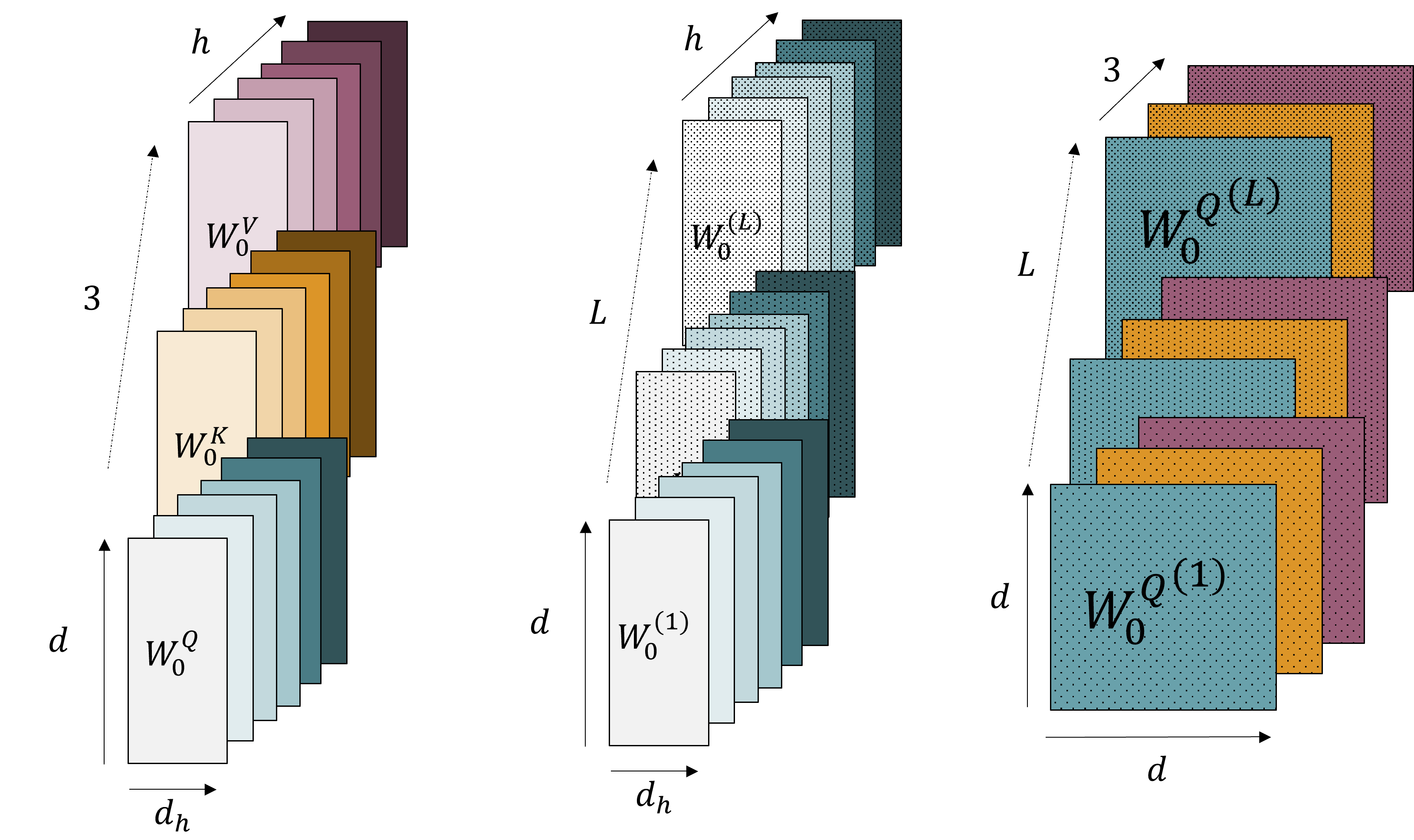}
        \caption{Combinations}
        \label{fig:combinations}
    \end{subfigure}
    \caption{TensLoRA representations for \textbf{Att}, \textbf{QKV}, \textbf{Depth}, and combinations of two of them (4D tensors).}
    \vspace{-5pt}
\end{figure*}

\subsection{LoRA}
The core idea behind LoRA is to adapt a model by adding learnable low-rank updates to its weight matrices (kept fixed). Formally, given a pre-trained weight matrix $W_0 \in \mathbb{R}^{d \times d}$ of inner dimension $d$, LoRA replaces it during training by:
\begin{equation}
    W = W_0 + \Delta W, \quad \text{with} \quad \Delta W = AB,
\end{equation}
where $A \in \mathbb{R}^{d \times r}$ and $B \in \mathbb{R}^{r \times d}$ are low-rank matrices, and $r \ll d$. Here, $W_0$ is frozen during adaptation, meaning that only matrices $A$ and $B$ can be trained. In a model with $n$ attention projections (per layer, from 1 to 4, among Query, Key, Value, and Output~\cite{vaswani2017attention}) and $L$ layers where LoRA is applied, the total number of LoRA matrices is $2nL$, leading to a total trainable parameter count of $2drnL$.

\subsection{Tensor Extensions}
In this work, we introduce seven tensor extensions of LoRA, each based on a unique way to combine the $2nL$ matrices into tensors. We explore all possible dimension combinations to create these extensions.

\begin{table*}[ht]
\caption{How existing methods fit into the proposed framework. *CaRA also merges \textbf{QKV} and \textbf{Depth} dimensions.}
\centering
\begin{tabular}{|l|c|c|c|c|c|c|c|}
\hline
       & \textbf{Att} & \textbf{QKV} & \textbf{Depth} & \textbf{Att\_QKV} & \textbf{Att\_Depth} & \textbf{QKV\_Depth }& \textbf{Att\_QKV\_Depth} \\
\hline
FacT~\cite{jie2023fact}   &      &      &      &       & & X    &      \\
LoTR~\cite{bershatsky2024lotr}   &      &      & X    &       & & X    &      \\
LoRTA~\cite{hounie2024lorta}  &      &      &      &       & &      & X    \\
CaRA~\cite{veeramachenenicanonical}  &      &      &      &       & &      & X*    \\
\hline
\end{tabular}
\label{table:comparison_existing}
\vspace{-5pt}
\end{table*}

\subsubsection{Intra-Attention modules --- \textbf{Att}}
Transformers use multi-head attention, where each attention layer is composed of $h$ heads. Each head typically operates on a projection of size $d_h = d/h$, \ie a subset of the inner dimension. In most LoRA implementations, a single low-rank adaptation is shared across all heads. In the \textbf{Att} variant, we propose to separate the LoRA adaptation across heads, based on the hypothesis that different heads may be adapted using similar adaptation subspaces. This hypothesis may be too strong, because some previous work has suggested that each attention head holds a specialized role~\cite{gandelsman2023interpreting}.

Practically, \textbf{Att} applies tensor factorization on a third-order tensor $\mathcal{W}_0~\in~\mathbb{R}^{d\times d_h\times h}$, which stacks the head-specific matrices along a new axis, as presented in Figure~\ref{fig:att}.


\subsubsection{Inter-Attention modules --- \textbf{QKV}}
Each attention block consists of four projections~\cite{vaswani2017attention}: Query $\left(W_0^Q\in \mathbb{R}^{d \times d}\right)$, Key $\left(W_0^K\in \mathbb{R}^{d \times d}\right)$, Value $\left(W_0^V\in \mathbb{R}^{d \times d}\right)$, and Output $\left(W_0^O\in \mathbb{R}^{d \times d}\right)$. In this variant, we concatenate the Query, Key, and Value matrices to form a tensor $\mathcal{W}_0~\in~\mathbb{R}^{d\times  d\times 3}$, as presented in Figure~\ref{fig:qkv} (using the notation introduced earlier, $n=3$). Hence, \textbf{QKV} is based on the hypothesis that Query, Key, and Value may be adapted using similar adaptation subspaces.

We exclude the Output projection under the hypothesis that Query, Key, and Value share more structural similarity since they form the multi-head attention mechanism, making them more suitable for joint adaptation. This formulation additionally enables further extensions such as combining with the head dimension, as described in Section~\ref{sec:tensor_comb}.

\subsubsection{Depth of the model --- \textbf{Depth}}
Stacking the same type of projection (\eg Query matrices) from each of the $L$ layers results in a tensor $\mathcal{W}_0~\in~\mathbb{R}^{d \times d \times L}$, as presented in Figure~\ref{fig:depth}. \textbf{Depth} is based on the hypothesis that different layers may be adapted using similar subspaces.

\subsubsection{Combinations}
\label{sec:tensor_comb}
The above dimensions can be combined to create higher-order tensors. For instance, \textbf{Att\_QKV}, that combines \textbf{Att} and \textbf{QKV}, results in a 4th-order tensor $\mathcal{W}_0~\in~\mathbb{R}^{d \times d_h \times h \times 3}$. We present the three different 4th-order tensors in Figure~\ref{fig:combinations}. The combination of all methods, \textbf{Att\_QKV\_Depth}, leads to a 5th-order tensor $\mathcal{W}_0 \in \mathbb{R}^{d \times d_h \times h \times 3 \times L}$. 

\subsubsection{Relations to Related Work}
Several recent approaches have explored tensor-based formulations of LoRA. FacT~\cite{jie2023fact} stacks attention and MLP layers into a $12L \times d \times d$ tensor and applies Tensor-Train~\cite{oseledets2011tensor} or Tucker~\cite{tucker1966some} factorization. LoTR~\cite{bershatsky2024lotr} targets Query and Value projections, comparing a single $2L \times d \times d$ tensor (aggregating both projections across depth) with two separate $L \times d \times d$ tensors, and applies Tucker-2 factorization. LoRTA~\cite{hounie2024lorta} constructs a 5th-order tensor over layers, heads, and projection types (using the four projection types), and applies CP factorization~\cite{harshman1970foundations, carroll1970analysis}. CaRA~\cite{veeramachenenicanonical} follows a similar CP-based tensor but additionally merges layer and projection-type dimensions, and uses a second tensor for MLP layers. 
Table~\ref{table:comparison_existing} presents how these methods fit into our paradigm.

\section{Tensor Factorization}
\label{sec:tensor_formalism}
After defining the tensor constructions, we turn to their parametrization, for which we employ tensor factorization to limit the parameter count.

Tensor factorizations extend matrix factorization techniques to multi-way arrays (tensors)~\cite{kolda2009tensor}, and are widely used in signal processing to model high-dimensional structures, primarily as dimensionality reduction techniques~\cite{cichocki2015tensor}. In this first work, we focus on the Tucker factorization~\cite{tucker1966some}, while other tensor factorizations (Tensor-Train~\cite{oseledets2011tensor} or CP~\cite{harshman1970foundations, carroll1970analysis}) could also be included in future work.

The Tucker factorization~\cite{tucker1966some} consists of factorizing a tensor into a smaller core tensor and a set of mode-specific factor matrices. It generalizes the idea of matrix factorization to higher-order tensors by decoupling interactions along each mode. This allows the factorization to capture complex, structured relationships between modes while providing control over the number of components in each dimension. When orthogonality constraints are imposed on the factor matrices, Tucker becomes an extension of matrix SVD known as the Higher-Order Singular Value Decomposition (HOSVD)~\cite{de2000multilinear}.

For a third-order tensor $\mathcal{X} \in \mathbb{R}^{a \times b \times c}$, the Tucker factorization is written as:
          \begin{equation}
            \mathcal{X} \approx \mathcal{G} \times_1 \mathbf{A} \times_2 \mathbf{B} \times_3 \mathbf{C}
           \end{equation}
where $\mathcal{G} \in \mathbb{R}^{r_1 \times r_2 \times r_3}$ is called the core tensor, $A~\in~\mathbb{R}^{a\times r_1}$, $B~\in~\mathbb{R}^{b\times r_2}, C~\in~\mathbb{R}^{c\times r_3}$ are the factor matrices along the first, second, and third modes, respectively. The operator~$\times_n$ denotes the mode-$n$ tensor-matrix product. The number of factor matrices is equal to the number of modes of the original tensor, and the core tensor has the same number of modes as the original tensor.

Unlike factorizations that impose a uniform low-rank constraint across all modes (\eg CP), the Tucker factorization allows each mode to have its own rank $r_i$. This added flexibility is particularly advantageous in the adapter setting, where different modes (\eg attention heads, type of attention projection, layer depth, \etc) may benefit from varying levels of adaptation capacity. Carefully fitting ranks may lead to a better trade-off between compactness and expressivity. 


\section{Experiments}
\label{sec:expes}

\begin{table*}[ht]
\setlength{\tabcolsep}{3pt}
\caption{Scores for the different conditions and the different datasets. Parameter counts do not include the final classifier. We highlighted in bold the TensLoRA results that exceed LoRA. Results for baselines are reported in the original publications. LoTR and LoRTA only apply adapters to queries and keys. 
*Both FacT~\cite{jie2023fact} and CaRA~\cite{veeramachenenicanonical} are trained using only 1.000 samples, while we use the whole training dataset in our experiments; hence, comparisons should be taken with caution.}
\centering
\begin{tabular}{lcc|c|cc|cc}
\toprule
 & \multirow{3}{*}{Configuration} & \multirow{3}{*}{Rank} & \# Params & \multicolumn{2}{c|}{Vision (ViT)} & \multicolumn{2}{c}{Language (RobERTa-base)} \\
 &  &  & (\% of LoRA) & \multicolumn{2}{c|}{Accuracy} & MCC & Accuracy \\
 &  &  &  & DTD & EuroSAT & CoLA & MRPC \\
\midrule
 & LoRA           & $4$ & $221$k & $73.13\pm0.41$ & $98.45\pm0.19$ & $62.52\pm1.90$ & $88.68 \pm 0.84$ \\
\midrule
\multirow{7}{*}{\rotatebox{90}{Isorank}} & Att  & $4$ & $124$k ($56$\%) & $72.87\pm0.27$ & $98.06\pm0.14$ & $61.28\pm1.24$ & $86.89 \pm 1.24$ \\
 & QKV   & $4$ & $75$k ($33.7$\%) & $72.54\pm0.38$ & $98.30\pm0.15$ & $58.76\pm0.75$ & $87.45 \pm 0.77$ \\
 & Depth   & $4$ & $19$k ($8.5$\%) & $72.25\pm0.61$ & $97.88\pm0.36$ & $60.03\pm1.19$ & $86.71 \pm 0.76$ \\
 & Att\_QKV      & $4$ & $44$k ($19.8$\%) & $72.73\pm0.43$ & $98.07\pm0.34$ & $59.01\pm1.64$ & $88.04 \pm 1.13$ \\
 & Att\_Depth      & $4$ & $11$k ($5$\%) & $71.73\pm0.39$ & $97.64\pm0.30$ & $58.76\pm0.74$ & $86.08 \pm 0.47$ \\
 & QKV\_Depth    & $4$ & $6$k ($2.9$\%) & $71.59\pm0.31$ & $97.83\pm0.19$ & $58.09\pm2.35$ & $86.47 \pm 1.59$ \\
 & Att\_QKV\_Depth &$4$ & $4$k ($2$\%) & $71.07\pm0.31$ & $97.71\pm0.31$ & $58.13\pm2.01$ & $85.24 \pm 0.56$ \\

\midrule
\multirow{7}{*}{\rotatebox{90}{Isoparameters}} & Att &  {\footnotesize $d$:$7$, $d_h$:$4$, $h$:$12$} & $220$k ($99.5$\%) & $\boldsymbol{73.28}\pm0.58$ & $98.28\pm0.10$ & $60.04\pm1.85$ & $87.35 \pm 0.14$ \\
&QKV   & {\footnotesize $d$:$11$, $d$:$11$, $n$:$3$} & $207$k ($93.7$\%) & $\boldsymbol{73.27}\pm0.73$ & $98.34\pm0.16$ & $61.12\pm1.63$ & $88.48 \pm 0.57$ \\
&Depth & {\footnotesize $d$:$37$, $d$:$37$, $L$:$12$} & $220$k ($99.6$\%) & $73.05\pm0.61$ & $\boldsymbol{98.48}\pm0.09$ & $62.17\pm1.27$ & $\boldsymbol{88.97}\pm0.77$ \\
&Att\_QKV      & {\footnotesize $d$:$16$, $d_h$:$9$, $h$:$12$, $n$:$3$} & $218$k ($98.7$\%) & $\boldsymbol{73.97}\pm0.57$ & $98.45\pm0.11$ & $\boldsymbol{63.22}\pm0.77$ & $88.29 \pm 0.66$ \\
&Att\_Depth     & {\footnotesize $d$:$23$, $d_h$:$16$, $h$:$12$, $L$:$12$} & $216$k ($97.6$\%) & $\boldsymbol{73.76}\pm0.57$ & $\boldsymbol{98.61}\pm0.07$ & $62.16\pm1.13$ & $88.48 \pm 0.81$ \\
&QKV\_Depth     & {\footnotesize $d$:$60$, $d$:$60$, $n$:$3$, $L$:$12$} & $222$k ($100.3$\%) & $\boldsymbol{74.13}\pm0.56$ & $\boldsymbol{98.62}\pm0.13$ & $\boldsymbol{62.74}\pm1.08$ & $\boldsymbol{88.92}\pm0.36$ \\
&Att\_QKV\_Depth & {\footnotesize $d$:$28$, $d_h$:$16$, $h$:$12$, $n$:$3$, $L$:$12$} & $216$k ($97.8$\%) & $\boldsymbol{73.36}\pm0.95$ & $\boldsymbol{98.57}\pm0.11$ & $\boldsymbol{62.86}\pm0.86$ & $\boldsymbol{89.07}\pm0.51$ \\
\midrule
 & FacT*~\cite{jie2023fact}  & $32$ & $\approx69$k & $70.8$ & $96.2$ & - & - \\
 & LoTR~\cite{bershatsky2024lotr}  & $88$ & $321$k & - & - & $61.3 \pm 0.6$  & $88.0 \pm 0.9$ \\
 & LoRTA~\cite{hounie2024lorta} & $16$ & $15$k & - & - & $64.32$ & $90.44$ \\
  & CaRA*~\cite{veeramachenenicanonical} & \small{$16$ (DTD), $32$ (EuroSAT)} & $\approx60$k & $71.8\pm0.16$ & $96.4\pm0.1$ & - & - \\ 
\bottomrule
\end{tabular}
\label{table:scores}
\vspace{-5pt}
\end{table*}

\subsection{Datasets}
We evaluated our methods on four datasets across two modalities (Vision and Language): DTD~\cite{cimpoi2014describing} and EuroSAT~\cite{helber2019eurosat} for Vision, and CoLA and MRPC for Language, both part of the GLUE benchmark~\cite{wang2018glue}. These datasets were used in related work (FacT~\cite{jie2023fact} and CaRA~\cite{veeramachenenicanonical} for Vision, and LoTR~\cite{bershatsky2024lotr} and LoRTA~\cite{hounie2024lorta} for Language). All these datasets are used in a classification setting.

\subsection{Experimental and Model details}
Experiments were conducted in PyTorch~\cite{paszke2019pytorch}, using two backbones downloaded from HuggingFace~\cite{transformersHF}: a Vision Transformer (ViT) pre-trained on ImageNet-21k~\cite{dosovitskiy2020image} for Vision, and RoBERTa~\cite{liu2019roberta} for Language. We used the AdamW optimizer with default values, and a Cosine Annealing scheduler for the learning rate, with peak value (after warmup) fixed to $1.10^{-3}$, and lowest value fixed to $1.10^{-6}$. Tensors were handled using Tensorly and Tensorly-Torch~\cite{kossaifi2019tensorly}.

During our experiments, we noticed that two parameters were largely impacting results: the scaling factor $\alpha$ (as presented in~\cite{bershatsky2024lotr}, also used in our implementation of LoRA) and the initialization of tensors. In this paper, based on preliminary experiments, we fixed $\alpha=4$ and the initialization of tensors to ``orthogonal,'' thereby resembling HOSVD~\cite{de2000multilinear}. 

\subsection{Rank Selection}
We consider two strategies for selecting the tensor ranks~$r_i$, representing two limit cases relative to LoRA: ``isorank'' and ``isoparameters''. In isorank, representing the lowest parameter count, all modes share the same rank, set equal to the LoRA rank. In isoparameters, ranks are chosen so that TensLoRA matches the number of parameters in LoRA, yielding an upper bound. Since multiple rank configurations could satisfy this constraint, we adopt a simple heuristic: the QKV dimension is fixed to rank 3 and Att and Depth to 12, corresponding exactly to the sizes of these modes. The two remaining ranks are then distributed so that modes of equal size share the same rank, and larger modes receive higher ranks, while keeping the total parameter count close to that of LoRA. We employ this simple heuristic to avoid overfitting ranks, aiming to isolate the impact of the tensor structure.

\subsection{Experimental results}
Experimental results are presented in Table~\ref{table:scores} where the LoRA rank $r$ is fixed to 4. We present results as the average and the standard deviation over five runs.

\textbf{QKV\_Depth} and \textbf{Att\_QKV\_Depth} consistently outperform LoRA under the isoparameters condition, establishing them as our most promising methods. This result validates the design choices made in related works, which employ similar tensor constructions. It also quantitatively confirms that leveraging tensors can improve LoRA performance with the same number of parameters. Interestingly, these are the only two conditions that combine all LoRA matrices into a single tensor rather than several, independent ones.

Under the isorank condition, no single technique surpasses the LoRA baseline. The compression may be too extreme, as four methods out of seven use less than 10\% of LoRA's parameters. A more in-depth ablation study would be needed to bridge the performance gap between the isorank and isoparameters conditions. It is noteworthy, though, that performance does not precisely follow the number of parameters, suggesting that some conditions are more promising.

In particular, the \textbf{Att} method proves to be the least effective, almost always yielding the worst results in our isoparameters tests. In isorank tests, the \textbf{QKV} method outperforms \textbf{Att} on both EuroSAT and MRPC datasets despite having fewer parameters. A similar trend is observed where \textbf{QKV\_Depth} outperforms \textbf{Att\_Depth}. These findings suggest that the QKV and depth dimensions are the most critical to group in tensors. This is likely due to higher redundancy, hence indicating that projections and layers share similar adaptation subspaces, whereas attention heads maintain specialized roles that degrade when forced into a shared representation. 

In addition, LoRTA~\cite{hounie2024lorta} outperforms our methods, suggesting that some of their parametrization is beneficial (in particular, adapters on the output projection).

\section{Conclusion}
In this work, we introduced \textit{TensLoRA}, a unified framework that generalizes low-rank adaptation into tensor-based formulations. By systematically aggregating LoRA updates into higher-order tensors and applying Tucker factorization, we highlighted a wide spectrum of possible adaptations that extend and encompass prior approaches. Our experiments show that adaptations do not perform equivalently, suggesting that redundancy across modes is not uniform. In particular, aggregating over the attention heads dimension appears less efficient than the other dimensions (attention projection and depth of the model). Although LoRA remains difficult to outperform with drastically fewer parameters, our analysis highlights promising tensor structures for future exploration, with some outperforming LoRA in similar parameter budgets. 

Future research should investigate more thorough rank allocation strategies, notably via ablation studies across different dimensions, explore alternative factorizations (CP, Tensor-Train), and extend the framework to other modules, such as output projections and MLP layers. Additionally, we suggest that TensLoRA may serve as a tool for interpretability, enabling the analysis of redundancies across modes and the identification of critical components in Transformers.

\newpage

\section{acknowledgments}
With the support of ANR JCJC ENDIVE ANR-24-CE23-7365 and ANR JCJC MusAIc ANR-25-CE23-5316. We also thank Arguilar Carlos, Makhlouf Ghaith, and Sridi Chadha for preliminary discussions and experiments.

\bibliographystyle{IEEEbib}
\bibliography{biblio}

\end{document}